\documentclass[runningheads]{llncs}
\usepackage[T1]{fontenc}
\usepackage{graphicx}
\usepackage{comment}
\usepackage{algorithm}
\usepackage{algpseudocode}
\usepackage{tikz}
\usepackage{soul}
\usepackage{color}    
\usepackage{amsmath}
\usepackage{graphicx}
\usepackage{subcaption}
\usepackage{makecell}
\usepackage{amsfonts}

\usetikzlibrary{arrows.meta, positioning, shapes.geometric, fit}
\begin{document}
\title{Risk-Aware Semantic Grounding for Trustworthy LLM-Based Robot Planning}
\titlerunning{Risk-Aware Semantic Grounding, LLM-Based Robot Planning}

\author{Łukasz Sobczak\inst{1}\orcidID{0000-0001-9439-1812} \and
Nur Keleşoğlu\inst{1}\orcidID{0000-0002-0306-7281} \and
Sławomir Piotr Nowak\inst{1}\orcidID{0000-0002-0775-4935}}
%\authorrunning{Ł. Sobczak et al.}

\institute{Institute of Theoretical and Applied Informatics, Polish Academy of Sciences, Gliwice 44-100, Poland 
\email{\{lsobczak,nkelesoglu,snowak\}@iitis.pl} }
% \and
% ABC Institute, Rupert-Karls-University %Heidelberg, Heidelberg, Germany\\

\maketitle              % typeset the header of the contribution
\begin{abstract}
Large language models (LLMs) are increasingly used as high-level planners in robot navigation, but their outputs may become unreliable when instructions are ambiguous, unsupported by the environment, or semantically inconsistent. This paper presents a Risk-Aware Semantic Grounding framework for trustworthy LLM-based robot planning. Unlike existing LLM-based planners that primarily optimize plan generation, we formulate semantic grounding reliability as a multi-dimensional risk estimation problem. The proposed architecture explicitly models grounding uncertainty through ambiguity, hallucination and semantic-conflict risks before planning occurs, enabling the system to decide whether to execute the instruction, request clarification, or reject it. To evaluate the approach, we introduce TRUST-NAV, a benchmark containing both standard navigation tasks and risk-inducing instruction scenarios. Experimental results show that while conventional LLM planners achieve strong performance on valid navigation tasks, the proposed framework substantially improves ambiguity detection and semantic conflict rejection. These findings suggest that trustworthy robot planning should be evaluated not only by task completion, but also by the ability to recognize when execution should not occur.

\end{abstract}

\section{Introduction}

Large Language Models (LLMs) have recently demonstrated remarkable capabilities in natural language understanding, reasoning, and task decomposition, motivating their adoption as high-level planners in autonomous robotic systems. Recent research has explored LLMs for language-guided navigation~\cite{anderson2018vision,shridhar2020alfred,tellex2011understanding}, embodied reasoning ~\cite{huang2022language}, task planning ~\cite{brohan2023can,pmlr-v205-huang23c,singh2023progprompt,liang2023code,driess2023palme}, and human-robot interaction, highlighting their potential to improve the flexibility and accessibility of robotic applications.

Despite these advances, LLM-based planning remains vulnerable to reliability, safety, and trustworthiness issues. Unlike traditional planners operating on explicitly defined symbolic representations, LLMs may generate syntactically plausible yet semantically incorrect outputs. Such failures include hallucinated objects, incorrect room assignments, unsupported assumptions about the environment, and ambiguous instruction grounding. In human-centered environments, these errors may reduce task performance and lead to unsafe behavior or loss of user trust.

Existing approaches ~\cite{ren2023knowno,liang2024introspective,yin2024safeagentbench,wang2025madra} improve planning reliability through semantic maps, tool-augmented reasoning, retrieval mechanisms, and structured environment representations. While these methods often improve task completion rates, they primarily focus on generating better plans rather than determining whether a plan should be generated at all. Consequently, many systems remain vulnerable when confronted with ambiguous instructions, references to nonexistent entities, or semantically inconsistent requests.

We argue that trustworthy robot planning should not be viewed solely as a plan-generation problem, but as a risk-aware semantic grounding problem in which the system first estimates grounding reliability and then decides whether planning is justified. This perspective shifts the objective from maximizing execution rates toward minimizing unsafe or semantically unsupported actions.

To address this challenge, we introduce a \textit{Risk-Aware Semantic Grounding Framework} for LLM-based robot planning. The framework models three grounding failures: ambiguity, hallucination, and semantic conflict. Based on these risk indicators, a dedicated decision layer determines whether an instruction should be executed, clarified, or rejected before planning begins. To systematically evaluate this capability, we also introduce TRUST-NAV, a benchmark for trustworthy language-guided robot planning under both standard and risk-inducing instruction scenarios.

The main contributions of this work are as follows:

\begin{enumerate}
\item We formulate trustworthy robot planning as a risk estimation problem and introduce the Risk-Aware Semantic Grounding Framework, which separates execution reliability assessment from plan generation using ambiguity, hallucination, and semantic-conflict analysis. 

\item  We introduce a risk-aware decision mechanism that determines whether a navigation instruction should be executed, clarified, or rejected before plan generation.

\item We develop TRUST-NAV, a benchmark specifically designed to evaluate trustworthy semantic navigation under ambiguity, hallucination, and semantic conflict scenarios. 

\item We show that explicit grounding-risk estimation improves ambiguity detection and semantic-conflict rejection while maintaining competitive planning performance.
\end{enumerate}

The remainder of this paper is organized as follows. Section~\ref{sec:related} reviews related work on LLM-based planning, semantic grounding, and trustworthy AI. Section~\ref{sec:methodology} presents the proposed framework. Section~\ref{sec:experimental_Setup} describes the experimental setup and TRUST-NAV. Section~\ref{sec:results} presents baseline models, evaluation metrics, and reports the experimental results. Finally, Section~\ref{sec:conclusion} concludes the paper and discusses limitations and future work.

\section{Related Works}
\label{sec:related}

This section reviews prior work in five areas: (1) semantic mapping for robot navigation, (2) language-guided navigation, (3) LLM-based planning in robotics, (4) trustworthy and uncertainty-aware LLM planning, and (5) decoupled planning architectures.

\subsection{Semantic Mapping for Robot Navigation}
Semantic mapping augments metric maps with object, region, and relational information. Early systems combined SLAM with object detection to label observed scenes~\cite{nuchter2008towards}. Later work focused on compact and structured representations that remain expressive while being efficient to maintain~\cite{sobczak2025visual}, as well as semantic organization that supports consistent indoor navigation~\cite{halama2025semantic}. More recent open-vocabulary scene graphs expose object-level spatial relations directly to language models for scene querying and planning~\cite{gu2024conceptgraphs}. These approaches treat the map as trustworthy once built, but they do not verify whether a new instruction is actually consistent with it.

\subsection{Language-Guided Navigation}
Instruction following has been studied extensively in simulation and on real robots. Benchmarks such as R2R and ALFRED require agents to execute multi-step instructions with sequential subgoals and object interactions~\cite{anderson2018vision,shridhar2020alfred}. One line of work maps language and perception directly to actions, which typically requires large training sets and is difficult to interpret. Another uses modular pipelines that translate instructions into symbolic goals or waypoints for classical planners~\cite{tellex2011understanding}. While more interpretable, these systems still assume that the instruction is valid and focus on task completion rather than on detecting whether execution is appropriate.

\subsection{LLM-Based Task Planning in Robotics}
LLMs have become popular as high-level planners that decompose complex tasks into executable steps. Early prompting studies showed that pre-trained models can produce usable zero-shot plans for embodied agents~\cite{huang2022language}. SayCan grounds the model in robot affordances by scoring candidate actions according to both goal relevance and feasibility~\cite{brohan2023can}. Inner Monologue incorporates environment and execution feedback into the prompt~\cite{pmlr-v205-huang23c}, while ProgPrompt and Code as Policies ask the model to generate structured programs or executable policy code~\cite{singh2023progprompt,liang2023code}. PaLM-E further integrates perception and language in a single model~\cite{driess2023palme}. These methods significantly improve flexibility, but they mostly assume that the instruction is well-posed and that referenced entities exist, rather than explicitly estimating whether grounding is reliable.

\subsection{Trustworthy and Uncertainty-Aware LLM Planning}
The reliability of LLM planners has itself become a research topic. KnowNo uses conformal prediction to calibrate uncertainty and ask for help when the prediction set remains ambiguous~\cite{ren2023knowno}. Introspective planning instead prompts the model to assess its own uncertainty and choose between acting and asking, reducing both unsafe and overly cautious behavior~\cite{liang2024introspective}. SafeAgentBench shows that embodied agents still often accept unsafe commands~\cite{yin2024safeagentbench}, while MADRA introduces a training-free debate module that evaluates instruction safety before execution~\cite{wang2025madra}.

We share this goal but formulate the problem differently. Rather than compressing grounding reliability into a single uncertainty score, we represent grounding risk through three interpretable signals: ambiguity, hallucination, and semantic conflict. The decision is therefore not only whether to act or ask, but whether to \emph{execute}, \emph{clarify}, or \emph{reject}. More importantly, we treat risk as a property of semantic grounding rather than as a property of the planner itself. This allows execution decisions to be made before planning begins and makes the reason for a rejection explicit. TRUST-NAV is built around these grounding failures rather than physical danger.

\subsection{Decoupled Planning Architectures}
Splitting high-level task planning from low-level motion control is a long-standing idea that improves robustness, modularity, and reuse~\cite{kaelbling2013integrated}. We extend this principle by inserting a risk-assessment stage between language understanding and motion execution. Because this stage is rule-based and does not require training, it can be tuned on a small validation set and allows us to isolate the contribution of risk gating from the planner itself and from the underlying map representation.

\section{Methodology}
\label{sec:methodology}

\begin{figure}[ht]
\centerline{\includegraphics[scale=0.205]{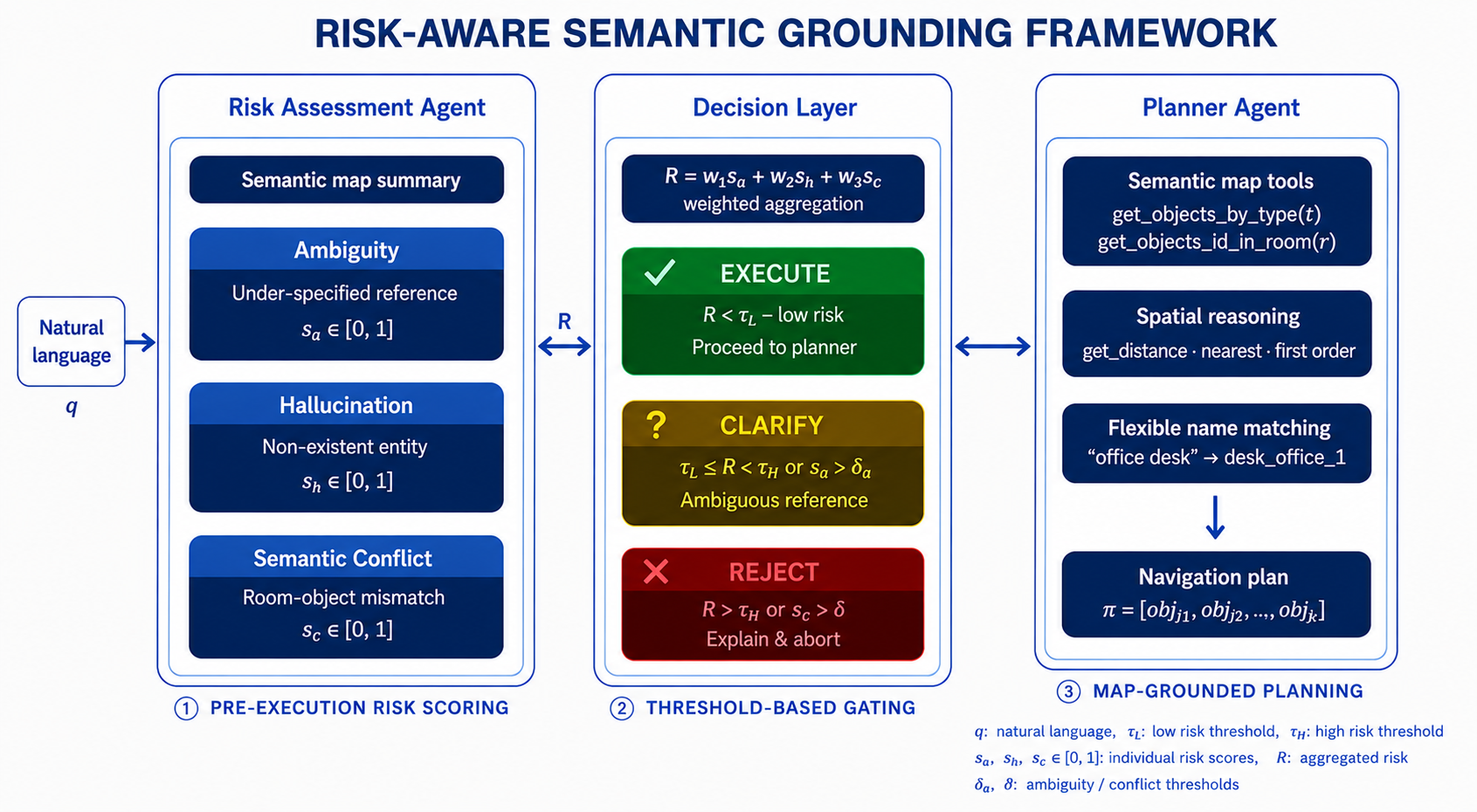}}
\caption{Architecture of the proposed Risk-Aware Semantic Grounding framework. Grounding risk is estimated prior to planning to support execute, clarify, or reject decisions.}
\label{fig:architecture}
\end{figure}

We propose a \textit{Risk-Aware Semantic Grounding Framework} (RA-SGF) that augments
conventional instruction-following pipelines for home-service robots with an
explicit, pre-execution safety layer. Rather than delegating ambiguity and
grounding errors entirely to the planning agent, our approach introduces two
dedicated components that act before any navigation plan is generated: a
\textit{Risk Assessment Agent} and a \textit{Decision Layer}. The overall
architecture is illustrated in Fig~\ref{fig:architecture}.

Formally, let $\mathcal{M} = (\mathcal{O}, \mathcal{R})$ denote the semantic
map of the environment, where $\mathcal{O}$ is the set of objects and
$\mathcal{R}$ is the set of rooms. Each object $o_i \in \mathcal{O}$ is
described by a tuple $(id_i,\, type_i,\, pos_i,\, props_i)$ encoding its
unique identifier, semantic type, 2-D position, and a set of key-value
properties. Each room $r_j \in \mathcal{R}$ is described by a tuple
$(id_j,\, type_j,\, O_j)$, where $O_j \subseteq \mathcal{O}$ is the set of
objects contained within room $r_j$. Given a natural-language instruction
$q$ and a starting position $p_0 \in \mathbb{R}^2$, the goal is to produce
an ordered sequence of object identifiers $\pi = [o_1, o_2, \ldots, o_n]$
constituting the navigation plan, or to request clarification / reject the
instruction when grounding is insufficiently reliable.

\subsection{Risk Assessment Agent}

The Risk Assessment Agent is an independent LLM agent that operates on the raw instruction $q$ and a compact textual summary of $\mathcal{M}$ injected into its system prompt. It produces three scalar risk scores, each representing a distinct failure mode in semantic grounding:

\begin{enumerate}
    \item \textbf{Ambiguity score} $s_a \in [0, 1]$: measures the degree to
    which $q$ is underspecified or admits multiple valid groundings without a
    discriminating spatial or semantic constraint. A score of $0$ indicates a
    unique, unambiguous referent; a score of $1$ indicates that the instruction
    cannot be grounded without further user input.

    \item \textbf{Hallucination score} $s_h \in [0, 1]$: measures the extent
    to which $q$ references objects, rooms, or properties that do not exist in
    $\mathcal{M}$. A score of $0$ indicates that all mentioned entities are
    present in the map; a score of $1$ indicates that the primary navigational
    target is entirely absent.

    \item \textbf{Semantic conflict score} $s_c \in [0, 1]$: measures the presence of logical contradictions between the instruction and the map
    topology, most commonly a room-object mismatch (e.g., referencing a
    refrigerator located in a bedroom when no such configuration exists in
    $\mathcal{M}$). A score of $0$ denotes full consistency; a score of $1$
    denotes a direct contradiction.
\end{enumerate}

The three scores are aggregated into a single \textit{overall risk score}
$R \in [0, 1]$ via a weighted linear combination:

\begin{equation}
    R = w_a \, s_a + w_h \, s_h + w_c \, s_c,
    \quad w_a + w_h + w_c = 1,
    \label{eq:risk_score}
\end{equation}

where $w_a$, $w_h$, and $w_c$ are non-negative scalar weights reflecting the
relative severity of each failure mode. In our experiments we set
$w_a = 0.30$, $w_h = 0.40$, and $w_c = 0.30$, assigning the highest weight
to hallucination as it represents the most unrecoverable failure: executing
a plan toward a non-existent target cannot be corrected at runtime without
replanning. The weights were selected empirically on a small validation subset and were not optimized on the test benchmark. %\hl{The weights were validated on a held-out calibration subset of the evaluation dataset; ablation experiments varying the weight configuration are reported in Section}~\ref{sec:ablation}.

\subsection{Decision Layer}

The Decision Layer is a deterministic, rule-based gate parameterized by two
soft thresholds $\tau_e$ and $\tau_c$ (with $\tau_e \leq \tau_c$) and three
hard per-dimension ceilings $\delta_a$, $\delta_h$, $\delta_c$:

\begin{equation}
    d(q) =
    \begin{cases}
        \texttt{reject}  & \text{if } s_h > \delta_h \text{ or } s_c > \delta_c, \\
        \texttt{clarify} & \text{if } s_a > \delta_a, \\
        \texttt{execute} & \text{if } R \leq \tau_e, \\
        \texttt{clarify} & \text{if } \tau_e < R \leq \tau_c, \\
        \texttt{reject}  & \text{if } R > \tau_c.
    \end{cases}
    \label{eq:decision}
\end{equation}

The hard-ceiling checks (first two cases) are evaluated before the soft
thresholds to prevent scenarios in which a high single-dimension risk is masked
by low scores in the remaining dimensions. For instance, an instruction with $s_h = 0.9$ but $s_a = s_c = 0.0$ yields $R = 0.36$ under our default weights, which falls within the clarification band $\tau_e < R \leq \tau_c$ and would therefore trigger only a request for clarification, even though the primary navigational target is entirely absent from $\mathcal{M}$; the hard ceiling $\delta_h$ ensures that such instructions
are rejected outright, regardless of the aggregate score. In our default configuration we use $\tau_e = 0.30$, $\tau_c = 0.70$, $\delta_a = 0.40$, $\delta_h = 0.80$, and $\delta_c = 0.85$.

The separation of hard and soft decision criteria provides two complementary
properties: \textit{precision} (hard ceilings handle clear-cut, unambiguous
failures with minimal false positives) and \textit{sensitivity} (soft
thresholds handle borderline cases where aggregate evidence suggests elevated
risk). This design also exposes interpretable, adjustable parameters that
practitioners can tune to match the risk tolerance of a specific deployment
environment.

\subsection{Planner Agent}

When the Decision Layer outputs \texttt{execute}, control is transferred to
the Planner Agent, which queries the semantic map through a set of tool
functions and generates an ordered navigation plan $\pi$. The Planner Agent
employs flexible name matching to resolve natural-language object descriptions
to map identifiers: given a mention such as \textit{``office desk''}, the
agent first retrieves the set of object identifiers in the office room via
\texttt{get\_objects\_id\_in\_room}, then intersects this set with the results
of \texttt{get\_objects\_by\_type("desk")}, and selects the unique matching
object. This two-stage lookup reduces the rate of spurious clarification
requests caused by surface-form mismatches between natural-language names and
map ontology labels.

When the instruction specifies multiple targets without an explicit ordering,
the agent resolves the visitation sequence using a nearest-first greedy
strategy. Given the robot's current position $p \in \mathbb{R}^2$ and a set
of candidate targets $T = \{o_1, \ldots, o_k\}$, the next waypoint is
selected as:

\begin{equation}
    o^* = \arg\min_{o_i \in T} \; \|pos_i - p\|_2,
    \label{eq:nearest}
\end{equation}

where $pos_i$ denotes the 2-D position of object $o_i$ and the search
iterates over the remaining unvisited targets until $T = \emptyset$.

The agent responds in a structured JSON format containing three fields:
(i) a \texttt{decision} field (\texttt{execute} / \texttt{clarify} /
\texttt{reject}) reflecting any residual uncertainty discovered during
grounding, (ii) a \texttt{plan} field containing the ordered list of object
identifiers, and (iii) an \texttt{answer} field providing a natural-language
confirmation or explanation for the operator.

% \subsection{Integration and Output}

The complete pipeline for a single instruction $q$ at starting position $p_0$ is summarized in Algorithm~\ref{alg:pipeline}.

\begin{algorithm}[ht]
\caption{Risk-Aware Semantic Grounding}
\label{alg:pipeline}
\begin{algorithmic}[1]
\Require instruction $q$, starting position $p_0$, semantic map $\mathcal{M}$,
         thresholds $\tau_e, \tau_c, \delta_a, \delta_h, \delta_c$, weights $w_a, w_h, w_c$
\Ensure  decision $d$, navigation plan $\pi$, response $\alpha$
\State $(s_a, s_h, s_c) \leftarrow \textsc{RiskAssessmentAgent}(q, \mathcal{M})$
\State $R \leftarrow w_a s_a + w_h s_h + w_c s_c$
\State $d \leftarrow \textsc{DecisionLayer}(s_a, s_h, s_c, R,\,
       \tau_e, \tau_c, \delta_a, \delta_h, \delta_c)$
\If{$d = \texttt{execute}$}
    \State $(\pi, \alpha) \leftarrow \textsc{PlannerAgent}(q, p_0, \mathcal{M})$
\ElsIf{$d = \texttt{clarify}$}
    \State $\pi \leftarrow [\,]$;\quad $\alpha \leftarrow \textsc{ClarificationMessage}(q, s_a)$
\Else
    \State $\pi \leftarrow [\,]$;\quad $\alpha \leftarrow \textsc{RejectionMessage}(q, s_h, s_c)$
\EndIf
\State \Return $(d, \pi, \alpha)$
\end{algorithmic}
\end{algorithm}

\section{Experimental Setup and Risk-Oriented Benchmark}
\label{sec:experimental_Setup}

\subsection{Experiment Environment: Smart Home Case}
\label{sec:simulation_environment}
\begin{figure}[htp]
\centerline{\includegraphics[scale=0.39]{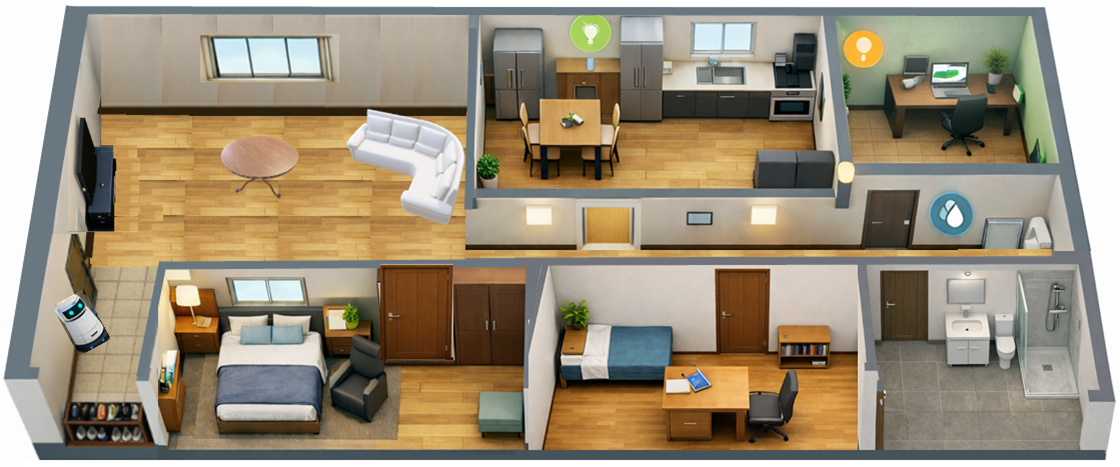}}
\caption{Smart home environment for a home assistant robot}
\label{fig:smart_home}
\end{figure}

The experimental environment is based on a single-floor apartment-style smart home designed to reflect a realistic residential setting, as illustrated in Fig.~\ref{fig:smart_home}. The total area of the apartment is $12 \times 8$ meters, corresponding to $96\,\mathrm{m}^2$. The environment is partitioned into eight semantically distinct regions, each representing a functional space commonly found in a household.

Specifically, the smart home comprises two bedrooms, a living room, a kitchen, an office, a bathroom, an entrance hall, and a corridor that connects the rooms. Each region is explicitly defined by fixed spatial boundaries and contains representative furniture and household objects (e.g., beds, desks, sofas, kitchen appliances) that serve as navigation and task targets in the experiments.

In the designed smart home environment, the mobile robot is assumed to start from a fixed and known initial position at coordinates $(x, y) = (0.2, 4.0)$ near the entrance door. This assumption simplifies the experimental setup by eliminating uncertainty in the robot’s initial localization and allows the evaluation to focus specifically on high-level planning and task execution performance.

\subsection{TRUST-NAV Benchmark}
\label{sec:trustnav}

To systematically evaluate the reliability and trustworthiness of LLM-based robotic planning systems, we introduce the \textbf{TRUST-NAV} (\textit{Trustworthy Navigation Benchmark}). It consists of five query categories representing different levels of planning complexity and risk. The first two categories evaluate standard planning capabilities, whereas the remaining categories focus on risk-aware decision making. The complete benchmark contains 206 natural-language instructions. The composition of the benchmark is summarized in Table~\ref{tab:trustnav}.

\begin{table}[t]
\centering
\caption{Composition of the TRUST-NAV benchmark.}
\label{tab:trustnav}
\begin{tabular}{l c @{\hspace{8mm}} c}
\hline
\textbf{Category} & \textbf{Number of Queries} & \textbf{Percentage (\%)} \\
\hline \hline
Single-Step Planning        & 34 & 16.50 \\
Multi-Step Planning         & 61 & 29.61 \\
Ambiguous Instructions      & 41 & 19.90 \\
Hallucination Scenarios     & 40 & 19.42 \\
Semantic Conflict Scenarios & 30 & 14.56 \\
\hline
\textbf{Total} & \textbf{206} & 100 \\
\hline
\end{tabular}
\end{table}

\begin{itemize}
\item \textbf{Single-Step Planning}: navigation instructions involving a single target object (e.g., \textit{Go to the sofa in the living room}). The expected behavior is to generate and execute a valid plan.

\item \textbf{Multi-Step Planning}: instructions containing ordered sequences of two to six navigation targets. These tasks evaluate task decomposition, sequential reasoning, and execution-order preservation.

\item \textbf{Ambiguous Instructions}: underspecified instructions that admit multiple valid groundings (e.g., \textit{Go to the chair}). A trustworthy system is expected to request clarification rather than arbitrarily select a target.

\item \textbf{Hallucination Scenarios}: instructions containing references to entities that do not exist in the environment (e.g., \textit{Go to the piano}). The expected behavior is to reject the instruction and report the inconsistency.

\item \textbf{Semantic Conflict Scenarios}: instructions whose referenced entities exist individually but form an invalid combination (e.g., a refrigerator in a bedroom). These tasks evaluate semantic consistency verification prior to planning.
\end{itemize}

The proposed benchmark therefore evaluates not only planning accuracy but also a system's ability to identify uncertainty, prevent hallucinations, and enforce semantic consistency. Such capabilities are essential for the deployment of trustworthy LLM-driven robotic systems operating in real-world human environments.

To support reproducibility and future research, the source code, benchmark, and evaluation scripts are publicly available at \url{https://github.com/iitis/Risk-Aware-Semantic-Grounding}.

\section{Results and Discussion}
\label{sec:results}

The proposed framework is evaluated from two complementary perspectives: planning performance and trustworthiness-aware decision making. While conventional navigation benchmarks focus primarily on task completion, trustworthy robotic systems must also recognize instructions that should be clarified or rejected before execution.

Accordingly, we first evaluate planning performance on executable navigation tasks and then assess trustworthiness-aware decision making using the metrics defined in Section~\ref{sec:evaluation_metrics}. All experiments were conducted using OpenAI \textbf{GPT-5.4-mini} as the underlying language model for all methods. The same model configuration was used across all evaluated methods to ensure a fair comparison.

\subsection{Compared Methods}
To evaluate the proposed model, we compare it against four baseline planners representing different levels of reasoning capability.

\subsubsection{Baseline 1: Nearest Semantic Object Planner (NSOP)}
The first baseline is a minimal semantic navigation strategy that reduces each instruction to a single navigation goal. It ignores instruction sequencing, object attributes, and intermediate constraints, and instead selects the nearest instance of the most relevant object type matched from the instruction tokens and the semantic map.

Formally, let $s \in \mathbb{R}^2$ denote the robot start position and let $\mathcal{O}_t = \{o_1, o_2, \dots, o_n\}$ be the set of objects of semantic type $t$. The selected goal is
\[
g^* = \arg\min_{o \in \mathcal{O}_t} \| s - p_o \|_2,
\]
where $p_o$ is the 2D position of object $o$. If no object type can be extracted, planning fails. This baseline serves as a lower-bound reference for a purely semantic, non-sequential, and geometry-light strategy.

\subsubsection{Baseline 2: Rule-Based Sequential Global Planner (RBSGP)}
The second baseline extends NSOP with deterministic instruction decomposition for multi-step navigation. It splits an instruction into sub-instructions using temporal and conjunctive cues such as ``then'', ``next'', ``after'', and punctuation, and processes each sub-instruction independently to extract an object type and, when available, an explicit object property.

If a matching object-property pair is found, the corresponding instance is selected; otherwise, the nearest object of the identified type is chosen using Euclidean distance. The final output is an ordered list of goal objects. While interpretable and deterministic, this baseline cannot detect ambiguity, hallucinated entities, or semantic inconsistencies.

\subsubsection{Baseline 3: Simple LLM Planner (SLLmP)}
The third baseline uses an LLM to directly decompose instructions into ordered navigation targets, but removes all geometric information from the map. The LLM operates only on object identities, types, and properties embedded in the prompt, without access to object coordinates or spatial distances.

Given an instruction $I$ and a simplified semantic map $\mathcal{M}_s$, the planner computes
\[
\Pi = \text{LLM}(I, \mathcal{M}_s),
\]
where $\Pi = [o_1, o_2, \dots, o_k]$ is an ordered list of object identifiers. No geometric reasoning or feasibility checking is performed. This baseline isolates the contribution of semantic reasoning by LLMs without explicit spatial grounding.

\subsubsection{Baseline 4: Tool-Augmented LLM Planner Agent (TA-LLmPA)}
The fourth baseline is a tool-augmented LLM planner that has access to the full geometric map and reasons iteratively through an agentic loop. At each step, it may call tools that retrieve object and room lists, filter objects, and compute distances. This enables it to resolve multi-step instructions, perform room-qualified lookups, and select spatially optimal visitation orders using Euclidean distance minimization (Eq.~\ref{eq:nearest}).

Formally, let $\mathcal{T}$ denote the tool set and $\mathcal{M}$ the full semantic map. The planner computes
\[
\Pi = \text{LLM}_{\mathcal{T}}(q, \mathcal{M}),
\]
where the subscript $\mathcal{T}$ indicates iterative tool access before producing the final output $\Pi = [o_1, o_2, \dots, o_k]$.

The agent is instructed to return a structured JSON response with a \texttt{decision} field (\texttt{execute} / \texttt{clarify} / \texttt{reject}), the navigation plan, and a natural-language explanation. Unlike the proposed method, it has no explicit risk quantification, per-dimension scoring, or threshold-based gating. This makes it the most direct comparator, since both systems use the same underlying LLM and identical tool access; any gain can therefore be attributed to the proposed risk-aware pre-execution layer.

\subsection{Evaluation Metrics}
\label{sec:evaluation_metrics}

The proposed framework is evaluated from both planning-performance and trustworthiness perspectives. To this end, we employ five complementary metrics measuring planning correctness, ambiguity recognition, hallucination rejection, semantic-conflict detection, and overall decision reliability.

\paragraph{Planning Accuracy (PA).}
PA measures the fraction of executable queries for which the generated plan exactly matches the ground-truth object sequence:
\begin{equation}
PA = \frac{C_{exec}}{N_{exec}}.
\end{equation}

\paragraph{Ambiguity Detection Rate (ADR).}
ADR measures the fraction of ambiguous instructions correctly classified as requiring clarification:
\begin{equation}
ADR = \frac{C_{amb}}{N_{amb}}.
\end{equation}

\paragraph{Hallucination Rejection Rate (HRR).}
HRR measures the fraction of hallucinated instructions correctly rejected:
\begin{equation}
HRR = \frac{C_{hall}}{N_{hall}}.
\end{equation}

\paragraph{Semantic Conflict Rejection Rate (SCR).}
SCR measures the fraction of semantically inconsistent instructions correctly rejected:
\begin{equation}
SCR = \frac{C_{conf}}{N_{conf}}.
\end{equation}

\paragraph{Decision Accuracy (DA).}
The proposed framework assigns each instruction to one of three outcomes:
\begin{equation}
% D \in {\texttt{execute}, \texttt{clarify}, \texttt{reject}}.
D \in \{\textit{execute}, \textit{clarify}, \textit{reject}\}
\end{equation}

Overall decision reliability is measured as
\begin{equation}
DA = \frac{C_{decision}}{N},
\end{equation}
where $N$ is the total number of benchmark queries and $C_{decision}$ is the number of correctly predicted decisions.

\subsection{Planning Performance}

\begin{figure*}[t]
\centering

\begin{subfigure}[t]{0.48\textwidth}
\centering
\includegraphics[width=\linewidth]{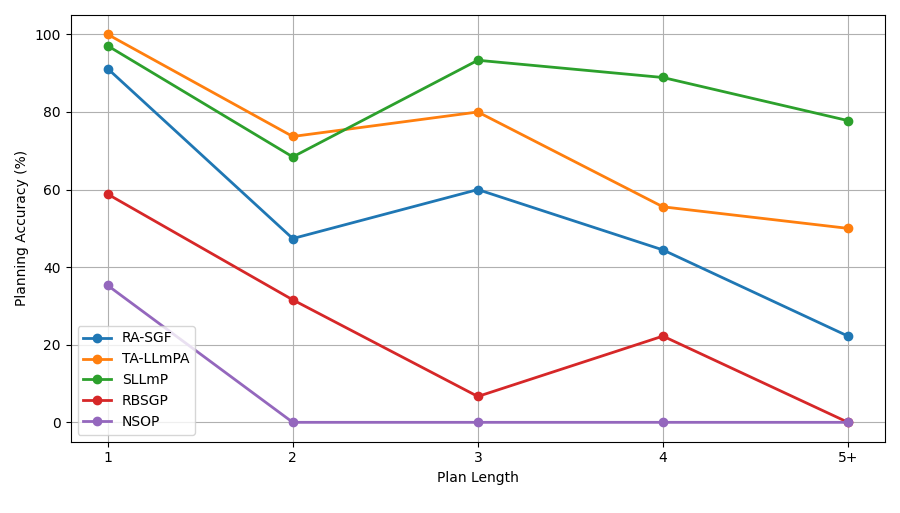}
\caption{Planning accuracy as a function of plan length.}
\label{fig:length_accuracy}
\end{subfigure}
\hfill
\begin{subfigure}[t]{0.48\textwidth}
\centering
\includegraphics[width=\linewidth]{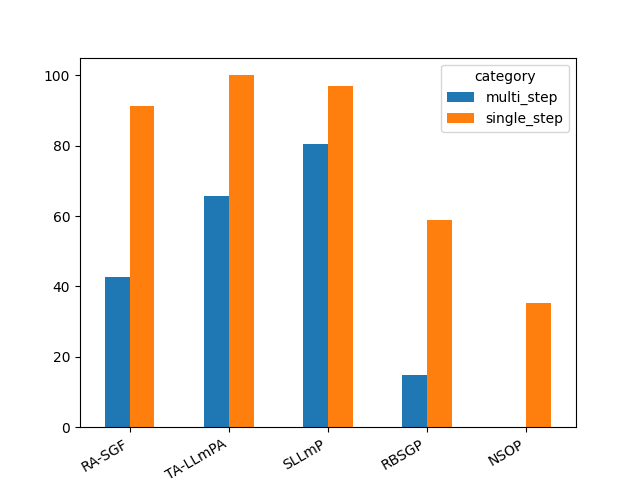}
\caption{Comparison of single-step and multi-step planning performance.}
\label{fig:single_multi}
\end{subfigure}
\caption{
Planning performance of all evaluated planners.%:
% (a) accuracy versus plan length;
% (b) accuracy on single-step and multi-step tasks.
}
\label{fig:planning_results}
\end{figure*}

We first evaluate planning performance on executable navigation tasks from TRUST-NAV, i.e., the Single-Step and Multi-Step categories. Table~\ref{tab:planning_accuracy} and Fig.~\ref{fig:planning_results} summarize the results.

As expected, planning accuracy decreases as task complexity increases. The strongest multi-step performance is achieved by SLLmP (80.33\%), while TA-LLmPA attains perfect single-step accuracy (100.00\%) but drops to 65.57\% on multi-step tasks. The proposed RA-SGF achieves 91.18\% and 42.62\%, respectively.

This reduction is expected, as the proposed framework prioritizes reliable pre-execution decision making over maximizing execution rates. By explicitly identifying instructions that should be clarified or rejected, the framework adopts a more conservative strategy than conventional planners. Consequently, planning accuracy alone does not fully capture system reliability, motivating the trustworthiness-oriented evaluation presented next.

\begin{table}[ht]
\centering
\caption{Overall planning accuracy (\%) on the TRUST-NAV benchmark.}
\label{tab:planning_accuracy}
\begin{tabular}{l @{\hspace{8mm}}c @{\hspace{8mm}} c}

\hline
\textbf{Planner} & \textbf{Single-Step} & \textbf{Multi-Step} \\
\hline \hline
Baseline 1: NSOP & 35.29 & 0.00 \\
Baseline 2: RBSGP & 58.82 & 14.75 \\
Baseline 3: SLLmP & 97.06 & 80.33 \\
Baseline 4: TA-LLmPA & 100.00 & 65.57 \\ 
Proposed Method: RA-SGF & 91.18 & 42.62 \\
\hline
\end{tabular}
\end{table}

\subsection{Evaluation under Risk-Inducing Scenarios}

TRUST-NAV also evaluates whether a planner can recognize instructions that should not be executed. In realistic deployments, robots may receive ambiguous, unsupported, or semantically inconsistent requests, where generating a plan is insufficient and the system must first decide whether execution is appropriate. To assess this capability, we measure decision accuracy across all benchmark categories.

\subsubsection{Decision Accuracy (DA)}

Table~\ref{tab:decision_accuracy} reports decision accuracy across all TRUST-NAV categories. Unlike planning accuracy, which measures whether the generated plan is correct, decision accuracy evaluates whether the system selects the appropriate high-level action (\textit{execute}, \textit{clarify}, or \textit{reject}).

The results show a clear distinction between raw planning performance and trustworthy decision making. SLLmP obtains the highest overall decision accuracy (84.95\%), largely because of strong performance on standard navigation tasks. The proposed RA-SGF is designed for more difficult cases and therefore prioritizes safe pre-execution decisions over maximizing overall execution rates. As a result, it does not achieve the highest score on every category, but it performs best on the two failure modes most closely related to trustworthy grounding.

\begin{table*}[ht]
\centering
\caption{Decision accuracy (\%) across different TRUST-NAV benchmark categories.}
\label{tab:decision_accuracy}
\begin{tabular}{l@{\hspace{2mm}}cccccc}
\hline
\textbf{Planner} &
\makecell{\textbf{Single}\\\textbf{Step}} &
\makecell{\textbf{Multi}\\\textbf{Step}} &
\makecell{\textbf{Ambi-}\\\textbf{guous}} &
\makecell{\textbf{Halluci-}\\\textbf{nation}} &
\makecell{\textbf{Semantic}\\\textbf{Conflict}} &
\textbf{Overall} \\
\hline \hline

Baseline 1: NSOP
& 55.88 & 65.57 & 0.00 & 82.50 & 13.33 & 46.60 \\ 

Baseline 2: RBSGP
& 85.29 & 100.00 & 0.00 & 77.50 & 13.33 & 60.68 \\ 

Baseline 3: SLLmP
& 97.06 & 93.44 & 78.05 & 67.50 & 86.67 & \textbf{84.95} \\ 

Baseline 4: TA-LLmPA
& \textbf{100.00} & 78.69 & 58.54 & \textbf{92.50} & 86.67 & 82.04 \\ 

\makecell{\textbf{Proposed Method: RA-SGF}}
& 91.18 & 72.13 & \textbf{87.80} & 82.50 & \textbf{96.67} & 83.98 \\

\hline
\end{tabular}
\end{table*}

\subsubsection{Ambiguity Detection Rate (ADR)}

The proposed RA-SGF achieves the highest ADR (87.80\%), outperforming SLLmP (78.05\%) and TA-LLmPA (58.54\%). This improvement comes from explicitly modeling ambiguity before plan generation, which allows the system to request clarification instead of committing to an arbitrary grounding. The result shows that ambiguity remains difficult for conventional LLM-based planners, even when they produce plausible plans.

\subsubsection{Hallucination Rejection Rate (HRR)}

For hallucination rejection, TA-LLmPA performs best (92.50\%), while the proposed method reaches 82.50\%. Even so, it still outperforms SLLmP (67.50\%) and reliably rejects unsupported references. This suggests that hallucinated entities are often easier to detect through direct tool access to the semantic map, so explicit risk modeling offers smaller gains here than in ambiguity or semantic conflict.

\subsubsection{Semantic Conflict Rejection Rate (SCR)}

The proposed method RA-SGF achieves the highest SCR (96.67\%), compared with 86.67\% for both TA-LLmPA and SLLmP. This result indicates that semantic conflicts require relational reasoning, not just entity verification, since all referenced objects may exist while their combination is still invalid. The dedicated semantic-conflict analysis in the Risk Assessment Agent therefore provides a substantial benefit in detecting logically inconsistent instructions before planning begins.

Overall, the ADR and SCR results support the main hypothesis of this work: explicit risk-aware semantic grounding is especially valuable when instructions are ambiguous or internally inconsistent, i.e., when generating an incorrect plan would be worse than declining execution.

\section{Conclusion}
\label{sec:conclusion}
This paper introduced a Risk-Aware Semantic Grounding Framework for trustworthy LLM-based robot planning. Unlike conventional planning architectures that focus primarily on generating executable navigation plans, the proposed framework explicitly evaluates the reliability of semantic grounding before execution through dedicated ambiguity, hallucination and semantic-conflict analysis.

To evaluate this capability, we introduced TRUST-NAV, a benchmark containing both standard navigation tasks and risk-inducing instruction scenarios. The benchmark enables systematic evaluation of not only planning performance but also trustworthiness-aware decision making.

Experimental results demonstrate an important distinction between planning accuracy and decision reliability. While conventional LLM planners achieve strong performance on valid navigation tasks, they remain vulnerable when confronted with ambiguous instructions, unsupported references, and semantically inconsistent requests. In contrast, the proposed framework achieves the highest ambiguity detection rate (ADR, 87.80\%) and semantic conflict rejection rate (SCR, 96.67\%), demonstrating that explicit risk assessment substantially improves robustness in scenarios where executing an incorrect plan may be more harmful than declining execution.

More broadly, our results suggest a shift in evaluation philosophy for LLM-based robotics: trustworthy systems should be assessed not only by how often they act correctly, but also by how reliably they recognize when action should not be taken. As LLM-based robots become increasingly integrated into human-centered environments, trustworthy decision-making mechanisms will be essential for ensuring safe and reliable operation.

\textbf{Limitations:} TRUST-NAV focuses on grounding-related failures in a static indoor environment and does not currently model perception errors, dynamic scene changes, or long-horizon embodied tasks. Furthermore, the benchmark is limited to a single semantic environment. As a result, the reported findings should be interpreted as evidence of the effectiveness of grounding-risk estimation rather than a complete evaluation of trustworthy embodied intelligence.

\textbf{Future Work:} Future work will investigate adaptive threshold calibration, richer semantic consistency models, larger and more diverse benchmark environments, and the integration of risk-aware grounding with real-world robotic platforms operating in dynamic environments.

\begin{credits}
\subsubsection{\ackname}
This work was partly supported by the National Centre for Research and Development (NCBR) and co-funded by the European Union under the European Funds for Modern Economy (FENG) Programme (SMART Konsorcja), project no.\ FENG.01.01-IP.01-A0GV/24-00, ``Advanced maintenance and diagnostic tool for IT start-ups''.

\subsubsection{\discintname}
The authors have no competing interests to declare that are relevant to the content of this article.
% It is now necessary to declare any competing interests or to specifically
% state that the authors have no competing interests. Please place the
% statement with a bold run-in heading in small font size beneath the
% (optional) acknowledgments\footnote{If EquinOCS, our proceedings submission
% system, is used, then the disclaimer can be provided directly in the system.},
% for example: The authors have no competing interests to declare that are
% relevant to the content of this article. Or: Author A has received research
% grants from Company W. Author B has received a speaker honorarium from
% Company X and owns stock in Company Y. Author C is a member of committee Z.

\end{credits}

%\noindent\textbf{Data and Code Availability.} The TRUST-NAV benchmark, the
%smart-home semantic map, all planner implementations and the evaluation scripts
%are available at \url{https://github.com/iitis/Risk-Aware-Semantic-Grounding}.


\bibliographystyle{IEEEtran}
\bibliography{./references_llm_nav}

\end{document}